\documentclass[sigconf]{acmart}
\AtBeginDocument{%
  }

\usepackage{enumitem}
\usepackage{multirow}
\usepackage{booktabs} 
\usepackage{xcolor, colortbl, soul, kotex}
\usepackage{ulem}
\usepackage{float}
\usepackage{algorithm, comment}
\usepackage{algorithmic}
\setlist[itemize]{itemsep=1pt, parsep=0pt, topsep=3pt}

\acmConference[WWW '26]{Proceedings of the ACM Web Conference 2026}{April 13--17, 2026}{Dubai, United Arab Emirates}

\renewcommand\footnotetextcopyrightpermission[1]{} 
\newcommand{\del}[1]{}

\begin{document}

\title{\textit{Suppression} or \textit{Deletion}: A Restoration-Based Representation-Level Analysis of Machine Unlearning}

\titlenote{Accepted at The Web Conference 2026 (WWW '26)}

\author{Yurim Jang}
\affiliation{%
  \department{Department of Artificial Intelligence}
  \institution{Sungkyunkwan University}
  \city{Suwon}
  \country{Republic of Korea}
}
\email{jyl8755@g.skku.edu}

\author{Jaeung Lee}
\affiliation{%
  \department{Department of Computer Science and Engineering}
  \institution{Sungkyunkwan University}
  \city{Suwon}
  \country{Republic of Korea}
}
\email{dlwodnd00@skku.edu}

\author{Dohyun Kim}
\affiliation{%
  \department{Department of Computer Science and Engineering}
  \institution{Sungkyunkwan University}
  \city{Suwon}
  \country{Republic of Korea}
}
\email{kimdoh0423@g.skku.edu}

\author{Jaemin Jo}
\affiliation{%
  \department{Department of Computer Science and Engineering}
  \institution{Sungkyunkwan University}
  \city{Suwon}
  \country{Republic of Korea}
}
\email{jmjo@skku.edu}

\author{Simon S. Woo}
\authornote{Corresponding author.}
\affiliation{%
  \department{Department of Computer Science and Engineering}
  \institution{Sungkyunkwan University}
  \city{Suwon}
  \country{Republic of Korea}
}
\email{swoo@g.skku.edu}

\renewcommand{\shortauthors}{Yurim Jang, Jaeung Lee, Dohyun Kim, Jaemin Jo, and Simon S. Woo}

\begin{abstract}
As pretrained models are increasingly shared on the web, ensuring that models can forget or delete sensitive, copyrighted, or private information upon request has become crucial. Machine unlearning has been proposed to address this challenge. However, current evaluations for unlearning methods rely on output-based metrics, which cannot verify whether information is completely \textit{deleted} or merely \textit{suppressed} at the representation level, where \textit{suppression} is insufficient for true unlearning. 
To address this gap, we propose a novel restoration-based analysis framework that uses Sparse Autoencoders to identify class-specific expert features in intermediate layers and applies inference-time steering to quantitatively distinguish between \textit{suppression} and \textit{deletion}. 
Applying our framework to 12 major unlearning methods in image classification tasks, 
we find that most methods achieve high restoration rates of unlearned information, indicating that they only \textit{suppress} information at the decision-boundary level, while preserving semantic features in intermediate representations. 
Notably, even retraining from pretrained checkpoints shows high restoration, revealing that robust semantic features inherited from pretraining are not removed by retraining.
These results demonstrate that representation-level retention poses significant risks overlooked by output-based metrics, highlighting the need for new unlearning evaluation criteria. 
We propose new evaluation guidelines that prioritize representation-level verification, especially for privacy-critical applications in the era of pre-trained models. Code available at \url{https://github.com/Yurim990507/suppression-or-deletion}
\end{abstract}



\maketitle

\section{Introduction}
The proliferation of model-sharing platforms such as Hugging Face~\cite{wolf2019huggingface} has democratized access to a wide range of useful pretrained models. Practitioners routinely download these models trained on large-scale web data, fine-tune them on proprietary datasets, and redistribute them within the web ecosystem. 
However, DNNs can memorize and leak sensitive personal information from web-scraped data~\cite{kim2023propile}. 
As regulatory frameworks such as the EU's GDPR~\cite{voigt2017eu} mandate the \textit{``right to be forgotten''}, machine unlearning (MU)~\cite{cao2015towards} has emerged to selectively remove such data influence.
A straightforward approach is to retrain the model from scratch using only the retained data. While this guarantees complete removal, it is often prohibitively expensive. Instead, most practical unlearning
methods adjust the model’s parameters to selectively remove the
influence of the target data without full retraining (i.e., approximate
unlearning).

Despite the development of various approximate unlearning methods, evaluations of their effectiveness remain insufficient and unclear, as they primarily rely on output-based metrics. 
For example, unlearning success is typically measured by accuracy on the forget set or by the performance of membership inference attacks (MIAs)~\cite{shokri2017membership}.
However, such metrics cannot confirm whether features of the unlearned data have been removed from the model's intermediate representations.
As a result, the risks of representation-level retention remain underexplored.
In this work, we differentiate between two possible outcomes of unlearning (\textit{deletion} vs. \textit{suppression}): \textit{deletion} refers to the complete removal of class-specific representations from all layers, while \textit{suppression} refers to the case where these representations remain encoded in intermediate layers but are masked at the output.
To fill this gap, we propose a novel restoration-based analysis framework that tests whether unlearned information can be recovered at the representation level. Using inference-time steering on features identified by Sparse Autoencoders (SAEs), we show that accuracy on the forget set can be restored, even for models that satisfy output-based evaluations.


Our contributions are as follows:
\begin{itemize}
    \item We introduce an analysis framework that uses SAEs to identify class-specific features and applies inference-time steering to distinguish \textit{suppression} from \textit{deletion}.
    \item We apply this framework to 12 major unlearning methods in image classification tasks and find that most methods merely \textit{suppress} rather than \textit{delete} class-specific representations.
    \item Based on our analysis, we propose design and evaluation guidelines for unlearning methods that emphasize representation-level verification over current output-based metrics.
\end{itemize}

\section{Background and Related Work}
\textbf{Machine Unlearning (MU)}~\cite{cao2015towards} aims to remove specific data influence from trained models. Various approximate unlearning methods have been proposed to avoid the computational cost of full retraining~\cite{chundawat2023can,kurmanji2023towards,fan2023salun,foster2024fast,jia2023model,tarun2023fast,goel2022towards}.
These methods are typically evaluated using output-based metrics such as accuracy on the forget set (lower is better) and retain set (higher is better), or using MIAs, which attempt to infer whether a sample was part of the training data from the model's outputs~\cite{shokri2017membership}.
Alternatively, research in~\cite{kim2025we, lee2025unlearning} compares unlearning methods to fully retrained models by examining layer-wise representational similarity.

However, the aforementioned approaches have critical limitations, since output-based evaluations cannot verify whether features are \textit{deleted} or merely \textit{suppressed}. 
While layer-wise comparisons with retrained models provide representation-level verification, they require retraining from scratch for each unlearning scenario, which is impractical for large-scale deployments.
Our work addresses these limitations by introducing a restoration-based analysis framework that tests whether features can be restored at the representation level without requiring retraining.

\noindent \textbf{Sparse Autoencoders}~\cite{olshausen1997sparse} are often used to identify interpretable features in model activations. 
They employ overcomplete representations ($m > d$), where $d$ is the input dimension and $m$ is the number of latent features, with sparsity constraints (e.g., TopK), encouraging sparse feature sets. 
SAEs are valuable for providing mechanistic interpretability~\cite{templeton2024scaling,stevens2025sparse}, as they isolate human-understandable concepts within neural networks.
This interpretability enables feature steering~\cite{templeton2024scaling}, modifying intermediate representations to control model behavior. 
Recent approaches manipulate SAE features by scaling them with a multiplier $\alpha$ to amplify or ablate their influence~\cite{templeton2024scaling,o2024steering}. 
We leverage SAEs to identify class-specific features and perform restoration experiments \del{to test whether unlearned features can be recovered }through steering, extending SAE-based interpretability to representation-level MU evaluation.


\section{\textit{Suppression} or \textit{Deletion} Framework}

\begin{figure}[t]
\centering
\includegraphics[width=\columnwidth]{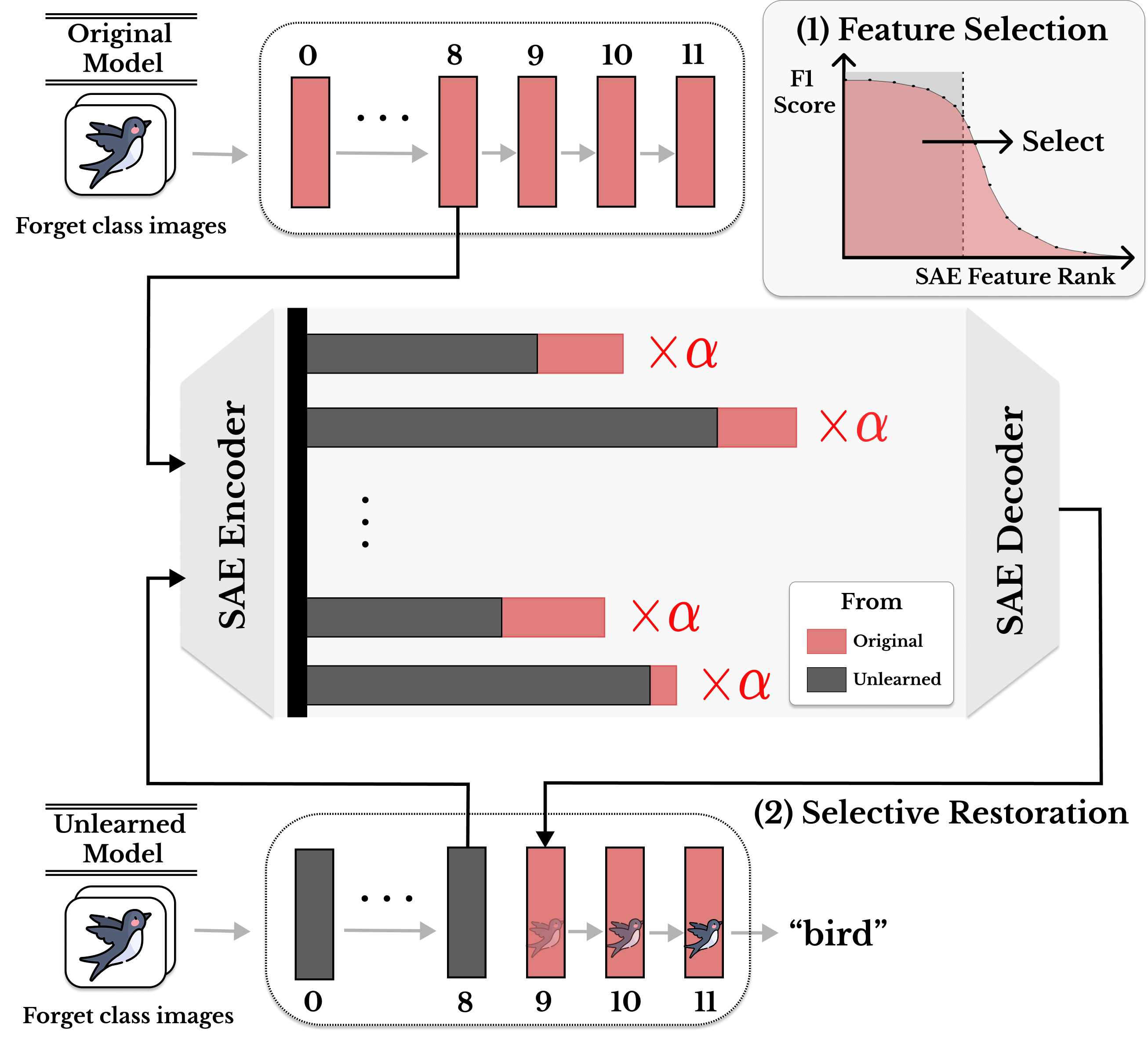}
\caption{Overview of the \textit{Suppression} or \textit{Deletion} framework. Top: Class-specific expert features are selected from SAE activations. Middle: Activations from the original and unlearned models are encoded via SAE and steered at these expert features to restore the representations. Bottom: The steered representations are decoded and propagated through the remaining layers; restoration of forget class predictions indicates \textit{suppression} rather than \textit{deletion}.
}
\label{fig:method}
\end{figure}

Prior work indicates that intermediate network layers tend to concentrate semantic information, while early layers capture low-level features and final layers are task-specific~\cite{joseph2025steering}.
Motivated by this observation, we focus our analysis on intermediate layers that serve as semantic bottlenecks. 
Our framework consists of two phases: (1) feature selection, and (2) selective restoration (see \hyperref[fig:method]{Figure~\ref{fig:method}}).

\begin{table}[t]
\centering
\caption{SAE ablation study on CIFAR-10 (Class 2, 20 features) and ImageNette (Class 7, 40 features). Ablating the expert features causes substantial forget class accuracy drops while maintaining retain class accuracy, validating class-specificity.}
\vspace{2pt}
\label{tab:sae_ablation}
{
\begin{tabular}{ll|ccc}
\hline\hline
\textbf{Dataset} & \textbf{Class} & \textbf{Layer 8} & \textbf{Layer 9} & \textbf{Layer 10} \\
\hline
\multirow{2}{*}{CIFAR-10} & Forget & -95.00 & -98.60 & -99.00 \\
& Retain & +0.10 & +0.18 & +0.24 \\
\hline
\multirow{2}{*}{ImageNette} & Forget & -82.80 & -96.44 & -98.80 \\
& Retain & +0.50 & +0.52 & +0.52 \\
\hline\hline
\end{tabular}
}
\vspace{6pt}
\end{table}

\textbf{Feature Selection.} We identify class-specific ``expert'' features through a four-step pipeline: 
(1) collect SAE activations from patch tokens at the $\ell$-th layer;
(2) filter out uninformative features that never activate or always activate, since such features do not contribute to class discrimination; 
(3) compute each feature's F1 score for class $c$ by calculating precision (i.e., probability of class $c$ when activated) and recall (i.e., probability of activation for class $c$);
(4) select the top $5K/4$ features per class by F1 score, where $K$ is the SAE's TopK sparsity level. 
To handle possible permutations of feature indices after unlearning, we align the features of the original and unlearned models using the Hungarian algorithm~\cite{kuhn1955hungarian}, yielding a matched feature set $\mathcal{F}_c$ for each class $c$.
We validate the class-specificity of these expert features (Table~\ref{tab:sae_ablation}): ablating them reduces forget class accuracy by more than 80\% across layers, while retain class accuracy changes within 1\%.

\begin{table*}[htbp]
\centering
\setlength{\tabcolsep}{4.5pt}
\caption{Our framework evaluates 12 major unlearning methods on CIFAR-10 and ImageNette, by measuring forget class accuracy before and after restoration across layers. \textcolor{red}{Red cells} indicate forget class accuracy over 80\% after restoration.}
\label{tab:restoration_results}
\resizebox{\textwidth}{!}{%
\begin{tabular}{l|c|c|c|c|c|c|c|c}
\hline
\multirow{3}{*}{\textbf{Method}} 
  & \multicolumn{4}{c|}{\textbf{CIFAR-10 (bird, Class 2)}} 
  & \multicolumn{4}{c}{\textbf{ImageNette (gas pump, Class 7)}} \\
\cline{2-9}
  & \multirow{2}{*}{\textbf{\shortstack{Unlearned\\Accuracy}}} 
  & \multicolumn{3}{c|}{\textbf{Restored Accuracy}} 
  & \multirow{2}{*}{\textbf{\shortstack{Unlearned\\Accuracy}}} 
  & \multicolumn{3}{c}{\textbf{Restored Accuracy}} \\
  \cline{3-5}\cline{7-9}
  &  & \textbf{Layer 8} & \textbf{Layer 9} & \textbf{Layer 10} 
     &  & \textbf{Layer 8} & \textbf{Layer 9} & \textbf{Layer 10} \\
\hline
Retrain & 0.00\% & \cellcolor{red!30}\textbf{95.70\%} (+95.70) & \cellcolor{red!30}\textbf{99.60\%} (+99.60) & \cellcolor{red!30}\textbf{99.40\%} (+99.40) & 0.00\% & 43.44\% (+43.44) & \cellcolor{red!30}\textbf{84.83\%} (+84.83) & \cellcolor{red!30}\textbf{98.97\%} (+98.97) \\
Finetune & 46.90\% & 45.90\% (-1.00) & 51.30\% (+4.40) & \cellcolor{red!30}\textbf{94.00\%} (+47.10) & 0.00\% & 0.72\% (+0.72) & 0.72\% (+0.72) & \cellcolor{red!30}\textbf{84.73\%} (+84.73) \\
AdvNegGrad & 0.00\% & \cellcolor{red!30}\textbf{100.00\%} (+100.00) & \cellcolor{red!30}\textbf{100.00\%} (+100.00) & \cellcolor{red!30}\textbf{100.00\%} (+100.00) & 0.00\% & 1.03\% (+1.03) & 21.08\% (+21.08) & \cellcolor{red!30}\textbf{82.01\%} (+82.01) \\
RandomLabel & 0.00\% & \cellcolor{red!30}\textbf{100.00\%} (+100.00) & \cellcolor{red!30}\textbf{100.00\%} (+100.00) & \cellcolor{red!30}\textbf{100.00\%} (+100.00) & 0.00\% & 4.30\% (+4.30) & \cellcolor{red!30}\textbf{82.82\%} (+82.82) & \cellcolor{red!30}\textbf{99.76\%} (+99.76) \\
Bad-T & 9.50\% & 0.50\% (-9.00) & 0.60\% (-8.90) & 0.00\% (-9.50) & 0.24\% & 2.86\% (+2.63) & 7.40\% (+7.16) & 1.19\% (+0.95) \\
SCRUB & 6.00\% & \cellcolor{red!30}\textbf{100.00\%} (+94.00) & \cellcolor{red!30}\textbf{100.00\%} (+94.00) & \cellcolor{red!30}\textbf{100.00\%} (+94.00) & 0.00\% & 0.26\% (+0.26) & 0.52\% (+0.52) & \cellcolor{red!30}\textbf{95.87\%} (+95.87) \\
SalUn & 0.00\% & 63.30\% (+63.30) & 72.80\% (+72.80) & \cellcolor{red!30}\textbf{96.30\%} (+96.30) & 0.00\% & 33.17\% (+33.17) & \cellcolor{red!30}\textbf{97.14\%} (+97.14) & \cellcolor{red!30}\textbf{100.00\%} (+100.00) \\
CF-K & 12.20\% & 0.10\% (-12.10) & 0.90\% (-11.30) & 27.20\% (+15.00) & 0.72\% & 0.00\% (-0.72) & 0.00\% (-0.72) & 0.24\% (-0.48) \\
EU-K & 0.00\% & 0.00\% (+0.00) & 0.00\% (+0.00) & 0.00\% (+0.00) & 0.00\% & 0.00\% (+0.00) & 0.00\% (+0.00) & 0.00\% (+0.00) \\
L1-Sparse & 0.00\% & \cellcolor{red!30}\textbf{97.60\%} (+97.60) & \cellcolor{red!30}\textbf{99.10\%} (+99.10) & \cellcolor{red!30}\textbf{99.40\%} (+99.40) & 0.00\% & \cellcolor{red!30}\textbf{84.73\%} (+84.73) & \cellcolor{red!30}\textbf{91.17\%} (+91.17) & \cellcolor{red!30}\textbf{96.90\%} (+96.90) \\
SSD & 1.30\% & 2.20\% (+0.90) & 26.40\% (+25.10) & 48.30\% (+47.00) & 0.00\% & 13.62\% (+13.62) & 21.59\% (+21.59) & 38.56\% (+38.56) \\
UNSIR & 0.00\% & 38.50\% (+38.50) & \cellcolor{red!30}\textbf{86.70\%} (+86.70) & \cellcolor{red!30}\textbf{86.70\%} (+86.70) & 0.00\% & 20.76\% (+20.76) & 51.55\% (+51.55) & \cellcolor{red!30}\textbf{82.34\%} (+82.34) \\
\hline
\end{tabular}
}
\end{table*}

\textbf{Selective Restoration.} For each target layer $\ell$, we extract the activation from that layer for both the original and unlearned models, denoted $h_{\text{orig}}$ and $h_{\text{unl}}$, respectively. 
After encoding these activations with the SAE, we selectively restore class $c$ features (i.e., expert features) of the original model in the unlearned model's representation.
Specifically, for each feature index $j \in \mathcal{F}_c$, we replace the
unlearned feature value with a weighted original value:
\begin{equation}
\hat{h}[j] = h_{\text{unl}}[j] + \alpha \big(h_{\text{orig}}[j] - h_{\text{unl}}[j]\big),
\end{equation}
where $\alpha$ is a steering coefficient. 
This yields a steered representation $\hat{h}$, in which the selected features match the original model.
We decode $\hat{h}$ through the SAE and feed it into the remaining layers of the unlearned model to produce an output. 
An increase in forget class accuracy after steering indicates that the unlearning only \textit{suppressed} the expert features for class $c$ rather than deleting them.

\section{Experiments}

\subsection{Experimental Setup}

\textbf{Datasets and Models.} We adopt ViT-B/16~\cite{dosovitskiy2020image} pretrained on ImageNet\nobreakdash-21K~\cite{wu2020visual} and fine-tune it on CIFAR-10~\cite{krizhevsky2009learning} and ImageNette~\cite{howard2019imagenette} (10-class subset of ImageNet).

\noindent \textbf{Unlearning Methods.} 
We evaluate 12 unlearning methods: Retrain (retraining the pretrained model on the retain set), Finetune, AdvNegGrad, RandomLabel, Bad-T~\cite{chundawat2023can}, SCRUB~\cite{kurmanji2023towards}, SalUn~\cite{fan2023salun}, CF-K/EU-K~\cite{goel2022towards}, L1-Sparse~\cite{jia2023model}, SSD~\cite{foster2024fast}, and UNSIR~\cite{tarun2023fast}.

\noindent \textbf{SAE Configuration.} We employ TopK SAEs with sparsity parameter $K=16$ for CIFAR-10 and $K=32$ for ImageNette.
For the restoration experiments, the steering multiplier is set to $\alpha=10$ to amplify the features.

\subsection{Results}
Applying our restoration-based framework to 12 unlearning methods, we find that the results in \hyperref[tab:restoration_results]{Table~\ref{tab:restoration_results}} reveal three main observations: the prevalence of \textit{suppression} over \textit{deletion}, the dependence of representation retention on layer depth and dataset complexity, and the method-level differences in achieving effective \textit{deletion}.

\textbf{Prevalence of \textit{Suppression} over \textit{Deletion}.}
\label{finding:finding1} 
We observe consistent \textit{suppression} patterns across most approximate unlearning methods, which cannot be measured by the existing output-based metrics. Despite achieving low (often 0\%) unlearned accuracy, these methods exhibit high restored accuracy.
In particular, methods that primarily adjust output mappings or loss functions (e.g., AdvNegGrad, SCRUB, RandomLabel) yield unlearned models that are restored to near-original accuracy in deeper layers of the network.
Similarly, other methods that perform parameter-level modifications, such as Finetune, SalUn, and UNSIR, also show high restorability of unlearned information. 
Notably, even Retrain shows high restored accuracy on both datasets. 
This indicates that robust semantic features inherited from pretraining persist. These methods merely \textit{suppress} class-specific representations without altering intermediate representations, falling short of \textit{deletion}.

\textbf{Effect of Layer Depth and Dataset Complexity.}
The effect of restoration is not uniform across layers but correlates with dataset complexity. 
Restoration peaks in middle layers (Layers~8--9) for the simpler CIFAR-10, but shifts to deeper layers (Layers~9--10) for the more complex ImageNette.
This pattern indicates that these layers act as ``semantic bottlenecks'', where class-specific information is most concentrated. The location of such bottlenecks is influenced by dataset complexity: simpler datasets (such as CIFAR\nobreakdash-10) concentrate features in middle layers, while more complex datasets (such as ImageNette) may push these bottlenecks to deeper layers.
Consequently, unlearning must be layer-aware and target the specific layers where critical information resides.

\textbf{Method-Level Differences in Achieving Effective \textit{Deletion}.} 
Our results enable the classification of unlearning methods based on their modification strategy. We distinguish two categories. 
The first category encompasses most approximate methods that rely on output or loss-level manipulation, as well as simple parameter adjustments such as Finetune. These methods are vulnerable to restoration, confirming they achieve only \textit{suppression}.
In contrast, methods in the other category perform targeted or structural modifications on intermediate layer parameters. 
In particular, EU-K, which employs layer reset, achieves 0\% restored accuracy across all layers, demonstrating effective \textit{deletion}. 
Similarly, methods using forms of weight dampening (e.g., SSD, Bad-T, CF-K) show significantly lower restoration rates.
This contrast demonstrates that achieving \textit{deletion} requires modifications that structurally alter or directly target the parameters of intermediate layers where semantic features are encoded.

\section{Guidelines for Future Unlearning}
Based on our experimental results, we propose the following guidelines for the design and evaluation of robust unlearning methods.

\textbf{Unlearning Method Design.}
Our observation that feature location depends on layer depth and dataset complexity implies that unlearning methods should be layer-aware. 
Effective unlearning design requires identifying and targeting semantic bottleneck layers where class-specific information concentrates, rather than applying uniform modifications. 
Furthermore, to achieve \textit{deletion}, methods must directly modify intermediate representations. 
Our analysis shows that methods relying on loss functions or output mappings are insufficient, as they leave representations intact.
Unlearning method designers must also address the persistence of pretrained knowledge. 
The high restored accuracy of Retrain demonstrates that representations encoded during pretraining are not removed by simple retraining.
This suggests that robust modifications, such as layer re-initialization (e.g., EU-K) or targeted parameter dampening, are necessary to remove deeply encoded semantic representations.

\textbf{Reliable Evaluation.}
Our experiments reveal a critical evaluation gap: unlearning methods can achieve 0\% forget accuracy yet remain fully restorable, demonstrating that conventional output-based metrics are insufficient for verifying true information removal. 
The widespread distribution of pretrained models in web ecosystems further amplifies these vulnerabilities.
As evidenced by Retrain's high restoration rates, robust semantic representations inherited from pretraining persist, thereby making such evaluations particularly misleading. 
Our framework provides a quantitative tool to distinguish \textit{suppression} from \textit{deletion} through restoration-based analysis.
We recommend that evaluations include layer-wise verification targeting critical semantic layers and restoration testing under minimal modifications.
For privacy-critical applications, representation-level auditing should be mandatory to ensure information is truly \textit{deleted}, not merely \textit{suppressed}.

\section{Limitations and Future Work}
\textbf{Generalizability.}
Our analysis focused on ViTs for image classification, and other architectures may distribute information differently, limiting direct generalizability.
Additionally, while class-wise unlearning served as an ideal setting for distinguishing suppression from deletion, extending this framework to other scenarios (e.g., instance-wise unlearning) remains a subject for future study.
A promising direction is to adapt expert feature selection to LLMs and generative models, as these architectures present unique challenges for defining and verifying information removal.

\textbf{SAE-based Interpretation.}
The features identified by SAEs depend on hyperparameters, such as sparsity level and expansion factor, and may not perfectly capture the model's complete internal behavior.
While our ablation study confirms class-specificity, caution is needed. Recent work has shown that SAEs can extract features even from randomly initialized models~\cite{heap2025sparse}, suggesting the need for alternative interpretability methods.

\section{Conclusion}
We introduce a restoration-based analysis framework using SAEs to quantitatively distinguish between representation \textit{deletion} and \textit{suppression}.
Applying this framework, we reveal that most approximate unlearning methods only \textit{suppress} information, leaving semantic representations persistent and recoverable in intermediate layers even when output-based metrics indicate successful unlearning.
This discrepancy poses significant risks, as models presumed safe are distributed via model-sharing platforms and redeployed, carrying exploitable persistent representations.
Based on our analysis, we propose key MU design guidelines: future evaluations must incorporate mechanistic verification, and effective unlearning requires layer-aware modifications that directly target intermediate representations.
By shifting evaluation from output behavior to internal mechanisms, our work provides a foundation for reliable privacy guarantees in the safe redistribution of pretrained models.

\begin{acks}
This work was partly supported by the Institute of Information \& Communications Technology Planning \& Evaluation (IITP) grants funded by the Korea government (MSIT) (RS-2022-II220688, RS-2019-II190421, RS-2024-00437849) and the National Research Foundation of Korea (NRF) grant funded by the Korea government (MSIT) (No. RS-2024-00356293).
\end{acks}

\bibliographystyle{abbrvnat}
{\small
\bibliography{ref}
}

\end{document}